\documentclass[runningheads]{llncs}
\usepackage{graphicx}
\usepackage{paralist}
\usepackage{tcolorbox}
\usepackage{multicol}
\usepackage{multirow}
\begin{document}
\title{An evaluation of template and ML-based generation of user-readable text from a knowledge graph}
\titlerunning{An evaluation of template and ML-based generation}
\author{Zola Mahlaza \and
C. Maria Keet \and
Jarryd Dunn \and
Matthew Poulter}
\authorrunning{Mahlaza et al.}
\institute{University of Cape Town, 18 University Avenue, Cape Town, South Africa
\email{\{zmahlaza,mkeet\}@cs.uct.ac.za}, \email{\{DNNJAR001,PLTMAT001\}@myuct.ac.za}}
\maketitle              % typeset the header of the contribution
\begin{abstract}
Typical user-friendly renderings of knowledge graphs are visualisations and natural language text. Within the latter HCI solution approach, data-driven natural language generation systems receive increased attention, but they are often outperformed by template-based systems due to suffering from errors such as content dropping, hallucination, or repetition. It is unknown which of those errors are  associated significantly with low quality judgements by humans who the text is aimed for, which hampers addressing errors based on their impact on improving human evaluations. We assessed their possible association with an experiment availing of expert and crowdsourced evaluations of human authored text, template generated text, and sequence-to-sequence model generated text. The results showed that there was no significant association between human authored texts with errors and the low human judgements of naturalness and quality. There was also no significant association between machine learning generated texts with dropped or hallucinated slots and the low human judgements of naturalness and quality. Thus, both approaches appear to be viable options for designing a natural language interface for knowledge graphs
\end{abstract}
\section{Introduction}
\label{sec:intro}

As the Google Knowledge Graph and related knowledge bases keep gaining popularity especially in industry, it elevates the importance of end-user interaction with the graph. The main tried and tested approaches are graphical visualisations and natural language text and, to some extent with the `infoboxes', tables as well. We focus on the natural langue interface for the interaction with knowledge graphs. While a template-based approach to generating natural language text has a long history, it demands resources, which has generated recent interest in data-driven systems for natural language generation (NLG) in the hope to reduce start-up costs. The data-driven approaches score relatively well on automated evaluation metrics, but humans evaluators---the ultimate consumers of the outputs---beg to differ: there were mixed results 
in the WebNLG\footnote{\url{https://webnlg-challenge.loria.fr/challenge\_2017/}} and E2E\footnote{\url{http://www.macs.hw.ac.uk/InteractionLab/E2E/}} NLG challenges held in 2017. For instance, 
a `grammar-infused' template system such as UPF-FORGe \cite{Mille2017} outperformed all the systems (except the baseline) in quality based on human judgements. Similarly, even though most template-based systems tend to have low scores for human judged quality and naturalness in the E2E challenge, the template-based systems TUDA \cite{DBLP:conf/inlg/PuzikovG18} and DANGNT \cite{Nguyen2017} outperform a large number of data-driven systems.

Puzikov and Gurevych \cite{DBLP:conf/inlg/PuzikovG18} and Smiley et al. \cite{DBLP:conf/inlg/SmileyDSS18} have offered possible reasons after analysing a small sample of the generated texts (100 and 25, respectively). They attribute the low quality of sequence-to-sequence (seq2seq) generated text to its grammaticality errors and the content dropping and hallucination habit (generating text not based on the graph's content) of such models. For instance, the graph may have \texttt{\{"Name\_ID":"Leo McKern",...\allowbreak "child": \{"mainsnak":"Abigail McKern"\}],...}, but due to the male-oriented training data (natural language text), this is then rendered as ``{\sf Leo McKern's son Abigail McKern}'', rather than ``child'' or, as it was more precisely in the original human authored text, ``daughter''. 
These issues may erode the trust by the user in the rendering of the graph and it should be at least minimised, and preferably avoided.
It is unknown, however, whether there is a significant association between low naturalness and quality judgement scores of the text and a model's amount of content dropping/hallucination or grammaticality errors.
An answer would enable a prioritization of the problems based on the impact their solutions would have on end users' perceived naturalness and quality.

In this paper, we investigate this association through the collection and analysis of expert and crowdsourced evaluations of human authored text, template generated text, and seq2seq model generated text. In particular, we seek to address the following research questions:
\begin{compactenum}
\item[RQ1] Are there significant differences in the perceived quality or naturalness of the text that is human authored, or generated by templates or data-driven methods?
\item[RQ2] Are texts with content dropping/hallucinations perceived by raters as having significantly lower naturalness and quality when compared to their counterparts?
\item[RQ3] Are texts with grammaticality errors perceived by raters as having significantly lower naturalness and quality when compared to their counterparts?
\end{compactenum}

Our results with the systems tested showed that there is {\em no significant difference} i) in quality and naturalness between human-authored, template-generated or seq2seq generated texts, ii) 
between human authored texts with errors and the low human judgements of naturalness and quality, and iii) 
between machine learning (ML) generated texts with dropped or hallucinated slots and the low human judgements of naturalness and quality.

The rest of the paper describes related work (Section~\ref{sec:relatedwork}), the experiment set-up  (Section~\ref{sec:methods}), presents the results (Section~\ref{sec:results}) that are discussed afterwards (Section~\ref{sec:discussion}) and concludes (Section~\ref{sec:conclusion}).

\section{Related work}
\label{sec:relatedwork}

Since we seek to compare approaches to  generating a natural language interface to knowledge graphs, the related works focuses on the extant literature on comparisons. The WebNLG-2017 challenge, with 9 participating systems, focused on the generation of text from RDF while the E2E NLG challenge, with 21 systems, focused on the same process but from other meaning representations. 
The ten best performing systems in the E2E challenge, based on the normalised average of automated metrics (BLEU, NIST, METEOR, ROUGE-L, and CIDEr), all use data-driven approaches. The best performing systems in the WebNLG challenge, based on automated metrics (BLUE, TER, and METEOR), are largely data-driven---the exception being the METEOR metric comparisons where a 
template system enhanced with some grammar (also called `grammar-infused' \cite{MK19mtsr})
outperforms all systems.

In both challenges, there are template-based systems that outperform a number of data-driven systems. For instance, the grammar-infused template system UPF-FORGe \cite{Mille2017} outperforms the other eight  systems of the WebNLG challenge (except the baseline) in quality based on human judgements.  Similarly, even though most template-based systems tend to have low scores for human judged quality and naturalness in the E2E challenge, the template-based systems TUDA \cite{DBLP:conf/inlg/PuzikovG18} and DANGNT \cite{Nguyen2017} outperform a large number of data-driven systems.

The only works that have investigated {\em why} template-based systems perform better than some data-driven models attributes the difference to the former having fewer grammatical errors, no content dropping and hallucination, and no degenerative repetitions \cite{DBLP:conf/inlg/PuzikovG18,DBLP:conf/inlg/SmileyDSS18}. While one can choose any of the problems to address (e.g., \cite{Holtzman2020} focus on degenerative repetition), from a human interaction viewpoint, a prioritization according to which one has a significant impact on improving human judgements of quality and naturalness may be most effective. To the best of our knowledge, there is no work that seeks to determine whether there is a significant association between the various mentioned errors and low human judgement errors.

\begin{figure}[h]
\centering
A

  \includegraphics[width=0.65\textwidth]{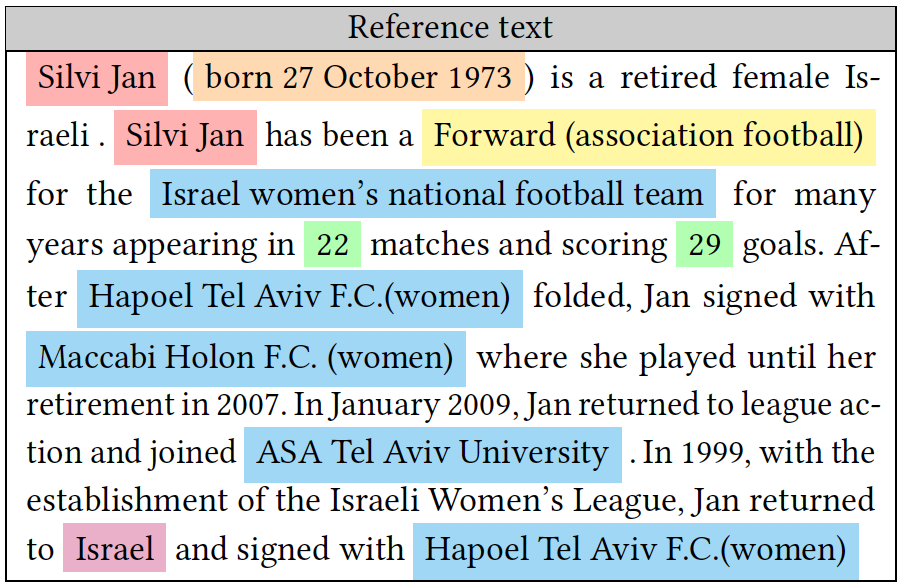}
  
  B
  
        \includegraphics[width=0.9\textwidth]{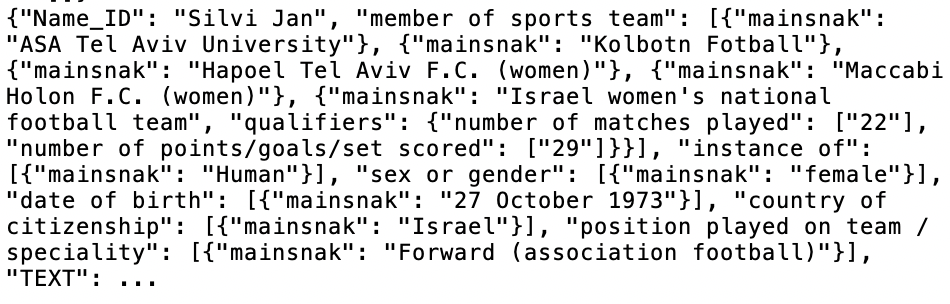}
    
  \label{fig:tempalteex}
  \caption{Dataset summary A: Original text from the Wikipedia people dataset; B: 
  the stored graph (structured data) that can be used for text generation (reference text at the ``...'' omitted).
(partially recreated from \cite{Wang2018})}
  \label{fig:wangdata}
\end{figure}

\section{Methods}
\label{sec:methods}

The aim of the experiment is to investigate whether there is a significant association between the various discussed errors and low human judgement errors and answer the research questions posed in Section~\ref{sec:intro}. We first describe the materials, being the data set with the knowledge graph used and the systems developed, and then the set-up of the human evaluation of the natural language texts generated.

\subsection{Materials}

We created two NLG systems, one template-based and the other data-driven, and used a dataset that is different from prior comparisons \cite{DBLP:conf/inlg/PuzikovG18,DBLP:conf/inlg/SmileyDSS18}.

\label{sec:systems}

\paragraph{Dataset}
We used an existing dataset that was extracted from Wikipedia and Wikidata \cite{Wang2018}. It is a collection of texts describing people and animals, but for this work we only considered the people subset. The subset contains exactly 100K persons, of which the general idea is shown in Fig.~\ref{fig:wangdata}; see \cite{Wang2018} for details. The dataset was split into three subsets: training (60\%), validation (30\%) and testing (10\%) and they were the same for both NLG systems.

\begin{figure}
 \centering
  \includegraphics[height=0.43\textheight]{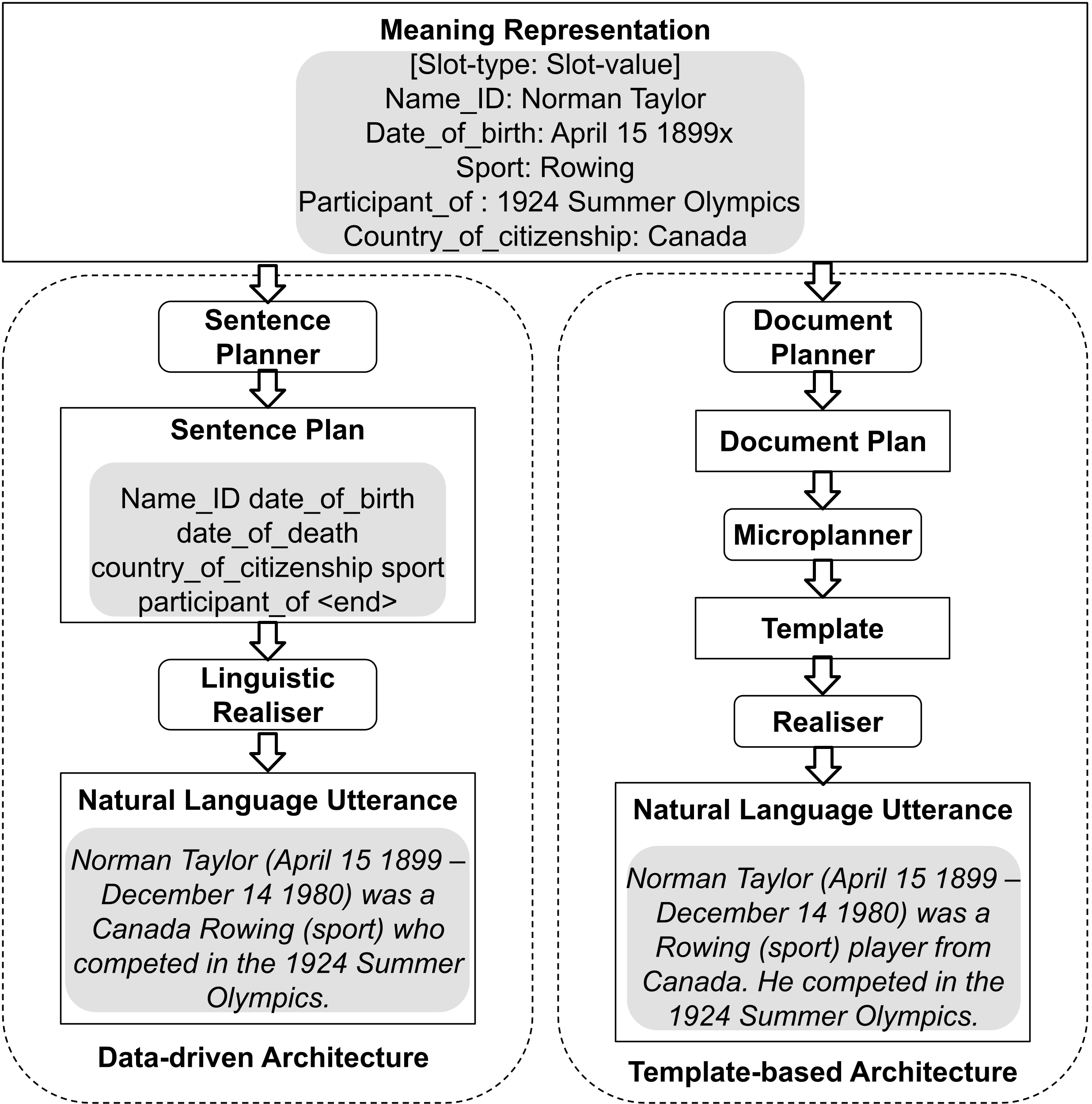}
  \caption{NLG pipelines of the two systems, with an example. ``Meaning representation'' at the top refers to the knowledge graph, with an example shown in abbreviated format.}
  \label{fig:summaryApproaches}
\end{figure}

\paragraph*{NLG systems} A summary of the approach of the two NLG systems is shown in Fig.~\ref{fig:summaryApproaches} with an example. 
    
  To build the template-based system, we used cosine similarity on the training dataset's input meaning representations and reference text to determine which sentences are similar and used them to manually extract templates. Due to the time-consuming nature of template creation, the templates only cover 75\% of the training data and thus, by design, lack full coverage (as distinct from unintended content dropping). A sampling of templates is shown in Fig.\ref{fig:tempalteex}, which also shows that templates with similar communicative goals for the same sequence of slot types are grouped together. In each of these template clusters, we further processed the templates to obtain subject-verb-object (SVO) underspecified trees. In each tree, there are three nodes and a tense annotation. The left-most node in a tree contains a template for generating the subject phrase, the right-most node contains template for generating the object phrase, and the node in between contains a verb. At generation time, the template-based system estimates the appropriate template cluster for each given input MR via its first module. After that, the cluster's SVO trees are ordered based on the input graph via the planning component and the realiser's first task is to insert slot values into the respective positions in each tree's templates. The last realisation task takes a tree, whose nodes are text, and uses SimpleNLG \cite{Gatt09} to inflect the middle node's verb based on the tense annotation and flattens the tree to obtain surface text.

\begin{figure}
  \centering
  \includegraphics[scale=0.45]{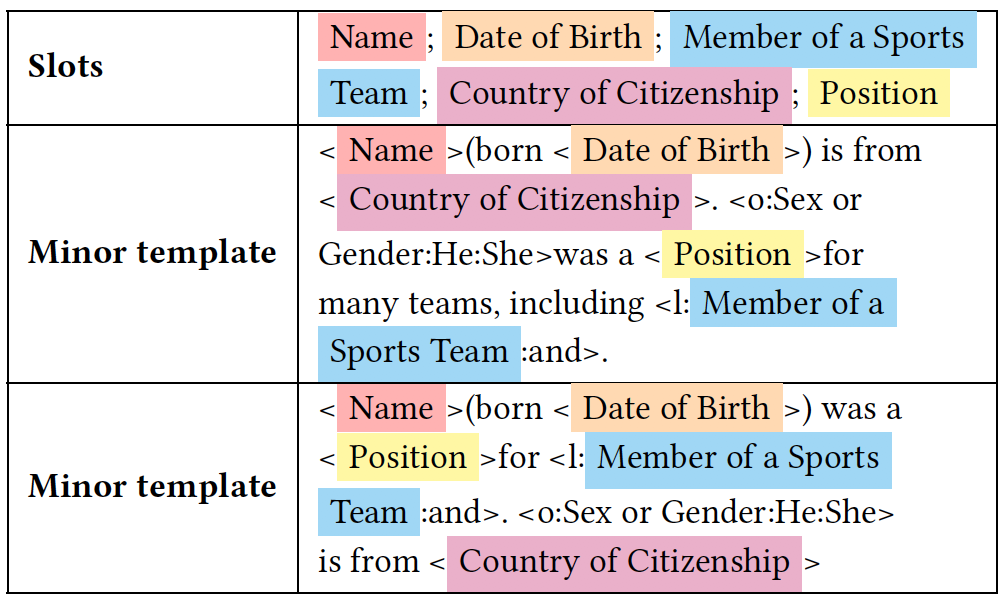}
  \caption{An example of templates used in the template-based system. The template do not necessarily include all variables available, which is largely due to resource constraints in crafting the templates.}
  \label{fig:tempalteex}
\end{figure}

The data-driven system consisted of a sentence planner and linguistic realiser. The sentence planner takes a set of tokens as input and tries to place these tokens into groups; these groups are then turned into natural language sentences by the linguistic realiser. The sentence planner is used to generate sentence plans using a Markov Chain type data structure where  each state consists of an n-gram of tokens and the transition probabilities are given by the probability of a new token coming after the current sequence of tokens. The realiser component has two components: the first is a seq2seq model created using OpenNMT \cite{Klein2017} for translating a sequence of ordered slot types to delexicalised text (i.e., templates) and the second component is responsible for inserting slot values into the respective positions, application of a simple rule to ensure that the output starts a capital letter, formatting whitespaces around parenthesis and punctuation, and removing special tokens produced by the seq2seq model (e.g., {\sf $\langle$unk$\rangle$} for out-of-vocabulary words). At runtime, once the sentence plans have been constructed, they are used as input to the realiser to produce a natural language sentence for each for the groups of tokens in the sentence plan.

A full `round trip' from the data to the systems' respective outputs is shown in Fig.~\ref{fig:fullcircle}. As can be seen, content is missed in the knowledge base creation, which may then be reduced further in the generated text; e.g., Fig.~\ref{fig:fullcircle}-D suffers both from content dropping (no place of death) and repetition (place of birth twice). A preliminary evaluation of the text was conducted with 91 respondents at the University of Cape Town. Each participant was presented with 10 paired descriptions taken from the two systems or a human reference text. This small sampling indicated that the template-based system was favoured, having sentences with higher clarity and fluency, and they were deemed more natural than those produced by the data-driven system. However, this is not a convincing picture due to the small number of texts evaluated. Details of the systems and results are available as supplementary material\footnote{\url{https://projects.cs.uct.ac.za/honsproj/cgi-bin/view/2019/dunn\_poulter.zip/results.html}}.

\begin{figure*}
\begin{tcolorbox} %[lowerbox=ignored]
%\begin{mdframed}

{
A. Original text that the tuple was created from: \\
\begin{footnotesize}
{\sf Dallas Green (baseball) ( August 4 1934 -- 2013 March 22 2017 ) was an American pitcher manager and executive in Major League Baseball. After playing for the Philadelphia Phillies Washington Senators and New York Mets from 1960 through 1967 he went on to manage the Phillies New York Yankees and Mets. Green was born in Newport, Delaware. After Green pitched to a 6 -- 2013 0 win-loss record and an 0.88 earned run average ( ERA ) in 1955 his junior year Jocko Collins a scout for the Philadelphia Phillies signed Green as an amateur free agent. Pitching for the Phillies Washington Senators and New York Mets Green had a career 20 -- 2013 22 record and 4.26 ERA in 185 total games with 46 games started. After acquiring left fielder Gary Matthews and center fielder Bob Dernier from Philadelphia before the 1984 season Green's Cubs became serious contenders for the first time in more than a decade. On March 22 2017 Green died at Hahnemann University Hospital in Philadelphia.}\\
\end{footnotesize}

B. Tuple in the knowledge base: 
\begin{small}
\begin{verbatim}
{"Name_ID": "Dallas Green (baseball)", 
"date of birth": [{"mainsnak": "August 4 1934"}], 
"instance of": [{"mainsnak": "Human"}], 
"sex or  gender": [{"mainsnak": "male"}], 
"member of sports team":  [{"mainsnak": "New York Mets"}, 
      {"mainsnak": "Philadelphia Phillies"}], 
"place of birth": [{"mainsnak": "Newport, Delaware"}], 
"date of death": [{"mainsnak": "March 22 2017"}], 
"place of death": [{"mainsnak": "Philadelphia"}] 
\end{verbatim}
\end{small}
}

\tcblower 

{\begin{multicols}{2}
C. Template-based text generated:\\
\begin{small}
Dallas Green (baseball) (August 4 1934 -- March 22 2017) was born in Newport, Delaware. He played for the New York Mets and Philadelphia Phillies. He died in Philadelphia. 
\end{small}

D. Data-driven text generated:\\
\begin{small}
Dallas Green (baseball) (August 4 1934 -- March 22 2017) was an American football position played on team / speciality who played for the New York Mets. He was born in Newport, Delaware in Newport, Delaware. 
\end{small}
\end{multicols}}
\end{tcolorbox}
  \caption{A: original text; B: corresponding tuple in the knowledge base of \cite{Wang2018}; C and D: text generated by our systems, which were evaluated on fluency, clarity, and naturalness.}
  \label{fig:fullcircle}
\end{figure*}
% more here: \url{https://projects.cs.uct.ac.za/honsproj/cgi-bin/view/2019/dunn_poulter.zip/results.html}

\subsection{Comprehensive evaluation procedure}

To gain a better understanding of the effects on a larger, and therewith more representative sample, we gathered 210 sentences (with their meaning representations, i.e., the graph snippets): 70 were human-authored, 70 were template-generated, and 70 were ML generated, using the systems as described in the previous section. We divided the sentences into five packages made up of combined human, ML, and template generated texts. The first four packages are made up of 45 sentences each and the last is made up of 30. Examples of the sentences are, among others:
\begin{small}
\begin{compactitem}
\item[T46]  Wilhelm D\"orpfeld (born on 26 December 1853 in Barmen and died on 25 April 1940 in Lefkada) was a Architect from Germany.
\item[ML46] Wilhelm D\"orpfeld (26 December 1853 -- 25 April 1940 in Lefkada) was a Germany Architect of Barmen and a member of the. He was buried at the Nydri in Nydri.
\item[T56]  Nathan Paulse (born 7 April 1982 in Cape Town) is from South Africa and he played for Ajax Cape Town F.C., Ajax Cape Town F.C., Hammarby IF, Ajax Cape Town F.C. and Ajax Cape Town F.C.
\item[ML56] Nathan Paulse (born 7 April 1982 in Cape Town) is a South Africa professional sport who plays for Ajax Cape Town F.C. Ajax Cape Town F.C. Hammarby IF and Ajax Cape Town F.C.. He made his debut for the Ajax Cape Town F.C. on 1 January 2010 in a 2 -- 2 win over Manchester City.
\end{compactitem}
\end{small}
See supplementary material at \url{https://github.com/AdeebNqo/TemplVsMLData} for all sentences evaluated and the appendix for illustrative pairs of graph snippets with the respective corresponding texts generated by the two systems. For each package, we added one question as an attention check, using the following text ``Please select \textbf{X} for Quality and \textbf{Y} for Naturalness. This is an instruction. It is not text to be evaluated." where \textbf{X} and \textbf{Y} are variables for values in the 5-point Likert scale 
labeled `very bad', `bad', `neutral', `good', and `very good'.

The five text packages were evaluated by humans who were remunerated 0.29 USD for each judgment on the crowdsourcing platform MTurk. Each participant was asked to rate the quality and naturalness of the text at most once. We ensured that for each question, there must be at least three judgements per text and at most 20. We presented texts to each participant and asked them to ``Rate the overall quality of the utterance, in terms of its grammatical correctness, fluency, adequacy and other important factors" on 
the 5-point Likert scale labelled `very bad ... very good' scale. 
We also asked them to rate whether ``[t]he utterance could have been produced by a native speaker" on 
the same scale. 
To ensure the quality of the responses, we eliminated participants who gave the same kind of judgment after the 10th text.

The faithfulness and grammaticality of the texts were  evaluated through analysis of the systems' internal representations and analysis of the text. We annotated and counted the number of dropouts and hallucinations per text. We also used LanguageTool\footnote{\url{https://github.com/languagetool-org/languagetool}} to find grammatical errors, manually analysed the resulting errors and classified them into specific types, and measured the number of error types per text.

For purposes of comparison, we categorised the evaluated texts based on the different kinds of errors present in them, 
assessed the judgements overall and compared to each other, including also squashing the 
judgements into a 3-value scale (positive, neutral, and negative) to assess tendencies, and to examine validity of converting to numerical values for statistical analysis. 

\section{Results}
\label{sec:results}

There were 50 respondents to the survey, 78\% of them self-reported as L1, 14\% as L2, and 8\% as L3 English speakers; 29 participants passed the attention check. No respondent was eliminated based on their responses, since none gave the same responses for all the texts after 10 judgements. Of the 70 sentences, 1-2 were lost during packaging, so there were 68 template generated (TT), 69 ML generated (TML), and 68 human authored (TH) text in the survey. Since the missing ones were not all the same, there were overall 66 graph snippets with texts for comparison of the texts across the modes. The results exclude ratings provided by participants who failed the attention check. The 210 texts and their 
judgements are given as supplementary material \url{https://github.com/AdeebNqo/TemplVsMLData}. 

\begin{table}
\centering
\caption{Average (avg.) judgement in each text category for quality and naturalness (natural.) when converted to a 5-point scale (1=Very bad, 5=Very good), and percentages of respective ratings over all rated sentences.}
\label{avgmedians}
\begin{tabular}{|l|l|l|l|l|l|l|} \hline
\textbf{Category} & \textbf{Avg. quality} & \textbf{very bad} & \textbf{bad} & \textbf{neutral} & \textbf{good} & \textbf{very good}  \\ 
&  \textbf{Avg. natural.} & \textbf{very bad} & \textbf{bad} & \textbf{neutral} & \textbf{good}	& \textbf{very good}  \\ \hline
\multirow{2}{*}{Human-authored (TH)}     & 3.54  & 0.7 & 12.6 & 32.9 & 39.5 & 14.3 \\ 
     & 3.59  &0.7 & 10.2 & 30.0 & 46.2 & 12.8  \\ \hline
\multirow{2}{*}{ML-generated (TML)}   & 3.64   &  0.5 & 11.2 & 32.2 & 35.6 & 20.5 \\ 
     & 3.60 & 1.2 & 8.6  & 28.0 & 51.2 & 11.0  \\ \hline
\multirow{2}{*}{Template-generated (TT)}     & {\bf 3.74} &  1.0 & 7.9 & 28.4 & {\bf 38.9} & {\bf 22.4} \\ 
     & {\bf 3.73}  & 1.0 & 6.0 & 27.8 & {\bf 48.3} & {\bf 15.8}  \\ \hline
\end{tabular}
\end{table}

Overall, there were 413 ratings for the human-authored sentences, 419 for the ML ones and 413 for the template-based ones; the percentages of ratings received is included in Table~\ref{avgmedians}, with template-based texts faring very slightly better when `good' and `very good' are combined into `positive'. Using the categorical squash and pitting positive against negative+natural, then all three are largely positive, with the NLG-generated text slightly beating negative+natural compared to TH for naturalness (52, 53, and 46  for the 66 graph snippets for TT, TML, and TH, respectively) and considerably outperforming TH text for quality (46, 40, and 31, respectively). For positive+neutral against negative, it reaches 66 (or 100\%) for TT and 65 for the other two, and 66 for all three for quality. Conversely, 
there are also minor differences regarding the ratio of texts with bad quality, with the TT and TH texts having 5 and 11 texts with $\geq 2$ people considering them as bad or very bad, and 9 of the TML texts. This also holds for naturalness, with 7 such negatively rated sentences for TT and TH texts and 8 of the TML texts.

Since the categorical data is intended to be roughly equidistant, we converted them to numerical data and calculated the respective individual and overall averages of the judgements, which are given in Table~\ref{avgmedians}. Using these to count `winners' for each comparable  text from the same graph snippet, the TT texts won 34 times, TML 17, and TH texts 18 times for quality and 34, 17, and 25 for naturalness, respectively (in case of a tie, both were awarded a point). Correlations between quality and naturalness a weak: 0.47 for TT, 0.45 for TML, and 0.51 for TH texts. Paired t-test on the numerical values for TT vs TML was significant for naturalness (p=0.0180) but not quality (p= 0.0879). That said, while the histograms for quality look roughly normally distributed, none of them are (Shapiro-Wilk tests), and naturalness is skewed to `good'/4, hence, these test outcomes are to be considered with caution.

The sample sentences of the previous section received an average quality of 3.6 for TT46 and 3.8 for its ML version (ML46), whereas TT56 received 4.3 template vs 3.7 for the ML56, whereas the sentences of Fig.~\ref{fig:fullcircle} (no. 3) both average to 4; for naturalness, their averages are, respectively: 3.9 vs 3.6, 3.9 vs 3.4, and 4.2 for both. Regarding positive vs neutral+negative for template vs ML: it is positive on quality for TT3, ML46, and both T56 and ML56, and for naturalness only TML3 was not positive and the rest positive.

\begin{figure}
\centering
\includegraphics[scale=0.5]{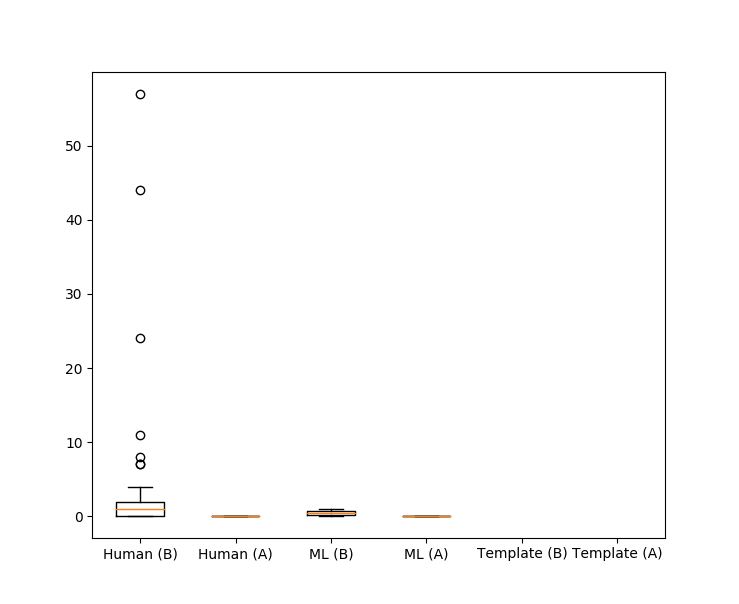}
\vspace{-2mm}
\caption{Number of grammar errors found by LanguageTool in the three categories of texts (B refers to number {\em before} and A refers to {\em after} manual verification.)}
\label{fig:grammarCounts}
\end{figure}

There were 9 types of valid errors detected by LanguageTool in the generated texts,  which were only present in the human authored and ML generated text;  the categories are shown in Table~\ref{tab:errorCategories}. The errors were only present in the human authored and machine learning generated text as shown in Figure~\ref{fig:grammarCounts}.

\begin{table}[t]
\small
\caption{List of error categories detected by LanguageTool, detected in the human and ML-generated text.}
\label{tab:errorCategories}
\begin{tabular}{|l|p{0.37\textwidth}|p{0.38\textwidth}|} \hline
{\bf Error category} & {\bf Description} & {\bf Example} \\ \hline \hline
PropOrthography & Orthographic errors resulting from the use of lower-case letters in proper nouns and acronyms, or upper-case for common nouns & ``gavra played with'' vs ``Gavra played with'' \\ \hline
Denonym & Errors due to incorrect form of the denonym & ``was an United States journalist'' vs ``was an American journalist'' \\ \hline
UnneccessarySpace & Introduction of space writing punctuation & ``he is a 7 ' 0 " 240 lb'' vs ``he is a 7'0" 240 lb'' \\ \hline
WrongSlotValue & Use of an incorrect value in a specific position, likely due to placing the incorrect value in a slot. & ``Nadine de rothschild (n\'{e}e Nadine de Rothschild'' vs. ``Nadine de Rothschild (n\'{e}e Nadine Lhopitalier'') \\ \hline
Agreement & Incorrect use of the indefinite article & ``is an United Kingdom" vs. ``is a United Kingdom'' \\ \hline
Typo & Typographical errors. & ``as an assistanr coach along" vs. ``as an assistant coach along'' \\ \hline
URLInfo & Errors due to the inclusion of text formatted for HTML's alt attribute & ``file : Fotothek df ps 0000106 Blick vom Turm des Neuen Rathauses.jpg'' \\ \hline
Repetition & Repetition of words or phrases & ``born in Belleville, New jersey New jersey and'' vs. ``born in Belleville, New Jersey and'') \\ \hline
MissingWordAfter & Missing word/phrases to complete a sentence & ``the United States Navy Lieutenant'' vs. ``the United States Navy as a Lieutenant''\\ \hline
\end{tabular}
\end{table}

\begin{figure}
\centering
\includegraphics[scale=0.60]{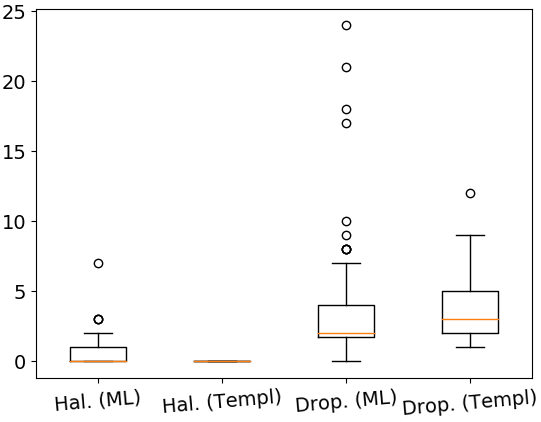}
\vspace{-2mm}
\caption{Number of hallucinated (Hal.) and dropped (Drop.) slots in the template-based (templ.) and data-driven (ML) NLG systems.}
\label{fig:slotCounts}
\end{figure}

Answering RQ1, then, 
there is a tendency that template-based texts receive more favourable ratings, but the effect is not unequivocal. Noting the caution of category to number conversion, it then does amount to a statistically significant difference between template and human for quality and between template and ML and human for naturalness.

Texts with content dropping/hallucinations are {\em not rated significantly lower} in naturalness and quality for the data-driven system, therewith answering RQ2 in the negative as well. 
In particular, the texts could have any of 4 categories of slot errors: (1) no hallucination, no dropping, (2) hallucination, dropping, (3) no hallucination, dropping, and (4) hallucination, no dropping. The template-based system's texts all are of category 3; hence, the question cannot be resolved. The ML model's text were present in the first three categories: there were 47 texts in category 3, 14 texts in category 2 and 7 texts in category 1. Regarding these ML-generated texts, there is no significant difference in perceived naturalness or quality between any of the error categories, based on Fisher's exact test (p=1).

Concerning RQ3, on grammar errors and ratings for naturalness and quality, we have observed again that there is no difference. More specifically, there are no detected errors in template-generated text and only 1 error in the ML generated texts; hence, the question cannot be resolved. In the human authored text, we observed that there is no significant difference in the ratio of good/bad quality and naturalness judgement between the texts with different kinds of errors (based on Fishers exact test (p = 1)). Likewise, there is no difference in the ratio of good/bad quality and naturalness judgements between the texts with errors when compared to those without, based on Fishers exact test (p = 1).

\section{Discussion}
\label{sec:discussion} 

Overall, thus, the evaluation could not settle the debate unequivocally either way in favour of template-based vs. data-driven approaches. Or, phrased in the positive: both approaches still seem viable options to a natural language interface for knowledge bases. This thus also suggest that the speculated attribution of reasons for any possible differences \cite{DBLP:conf/inlg/PuzikovG18,DBLP:conf/inlg/SmileyDSS18}, as noted in Section~\ref{sec:relatedwork}, turned out to not affect the human judgements of the sentences.

Some of the errors we have identified in the text belong to the error classes identified by Puzikov and Gurevych \cite{DBLP:conf/inlg/PuzikovG18}. Our Agreement, Typo, Repetition, and MissingWordAfter error categories all subsume their {\em bad grammar} error type. Their {\em modified contents} error type encapsulates WrongSlotValue, their {\em dropped contents} is the same as the dropped slot, their {\em punctuation errors} type encapsulates UnneccessarySpace, and their {\em questionable lexicalization} error type encapsulates Denonym errors. However, we have shown that the texts with the errors in Table~\ref{tab:errorCategories} do not have significantly worse human judgements than the texts without errors. Consequently, for well-resourced languages, a data-driven approach might thus still produce acceptable results. An issue still could be the hallucinations that an end-user consumer of the knowledge graph may not be aware of since they are not privy to the graph's actual content. This unless it is obvious, like assigning Abigail as being a ``son'', as noted in the introduction, which was a ML-generated sentence in the test set. Such potential bias in the language model, learned from the source data, is not an issue for a manually-crafted template-based systems. A template-based system seems also to be a route taken for the Abstract Wikipedia \cite{Vrandecic18}, although presumably soon `grammar infused' templates or fully grammar-based systems, like SimpleNLG \cite{Gatt09}, will be made possible. 

A grammar-supported approach would certainly be needed for the grammatically rich Niger-Congo B languages spoken in Sub-Saharan Africa \cite{KK16lre}. In that regard, `dropped content' may be a preferred limitation, especially for under-resourced languages. Considering Abstract Wikipedia's basis, Wikidata, the translations for properties are patchy, except for Afrikaans that has all relevant properties translated, but it should be possible to generate at least some basic person text in local languages. 
Straight-forward data points and short sentences may help populating currently non-existing pages. Several template and grammar-based text generation algorithms for knowledge graphs exist for Afrikaans and isiZulu \cite{KK16lre,STK16}, with translations and algorithms yet to be developed for the eight other official South African languages. 

\section{Conclusion}
\label{sec:conclusion}

The paper presented the first attempt to determine whether there is a significant association between texts with content dropping, hallucination, or grammatical errors and low human judgements of quality and naturalness. The two in-house developed NLG systems tested on the knowledge base of the Wiki people dataset revealed that: i) human authored texts are not significantly associated with low human judgements of naturalness and quality; and ii) the machine learning generated texts with dropped or hallucinated slots were also not significantly associated with low human judgements of naturalness and quality. Consequently, addressing one of these errors will not necessarily result in a significant improvement in perceived naturalness and quality and both approaches thus still could be used to generate natural language interfaces to knowledge graphs.

\subsection*{Acknowledgments}
This work was financially supported by Hasso Plattner Institute for Digital Engineering through the HPI Research School at UCT and the National Research Foundation (NRF) of South Africa (Grant Number 120852).
\bibliographystyle{splncs04}
\bibliography{semanticsrefs}

\section*{Appendix}
\begin{small}
Additional examples of pairs of graph snippets and the automatically generated text from them

\begin{description}
\item[Graph Snippet 7]  
\begin{verbatim}
{"Name_ID": "Ted Kleinhans", 
"date of death":  [{"mainsnak": "July 24 1985"}], 
"date of birth": [{"mainsnak": "April 8 1899"}], 
"instance of": [{"mainsnak": "Human"}], 
"sex or gender": [{"mainsnak": "male"}], 
"member of sports team": [{"mainsnak": "Philadelphia Phillies"}, 
{"mainsnak": "Cincinnati Reds"}, 
{"mainsnak": "New York Yankees"}], 
"place of birth": [{"mainsnak": "Deer Park, Wisconsin"}], 
"place of death": [{"mainsnak": "Redington Beach, Florida"}] 
\end{verbatim} 

\item[TML7:] Ted Kleinhans (April 8 1899 in Deer Park, Wisconsin -- July 24 1985 in Redington Beach, Florida) was a Major League Baseball pitcher who played for the Philadelphia Phillies Cincinnati Reds and New York Yankees. 

\item[TT7:] Ted Kleinhans (April 8 1899 -- July 24 1985) was born in Deer Park, Wisconsin. He played for the Philadelphia Phillies, Cincinnati Reds and New York Yankees. He died in Redington Beach, Florida.
\end{description}

\begin{description}
\item[Graph Snippet 59]  
\begin{verbatim}
{"Name_ID": "Souleymane Nd\u00e9n\u00e9 Ndiaye", 
"sex or gender": [{"mainsnak": "male"}], 
"date of birth": [{"mainsnak": "6 August 1958"}], 
"member of political party": [{"mainsnak": "Senegalese Democratic 
Party"}], 
"country of citizenship": [{"mainsnak": "Senegal"}], 
"instance of": [{"mainsnak": "Human"}], 
"place of birth": [{"mainsnak": "Kaolack"}], 
"occupation": [{"mainsnak": "Politician"}, {"mainsnak": "Lawyer"}] \end{verbatim}

\item[TML59:]   Souleymane Nd\'en\'e Ndiaye (born 6 August 1958 in Kaolack) is a Senegal Politician and Lawyer of the Senegalese Democratic Party.

\item[TT59:]  Souleymane Nd\'en\'e Ndiaye (born 6 August 1958 in Kaolack, Senegal) is a Politician and Lawyer.

\end{description}

\begin{description}
\item[Graph Snippet 52]  
\begin{verbatim}
{"Name_ID": "Xu Huaiwen", 
"sex or gender": [{"mainsnak": "female"}], 
"date of birth": [{"mainsnak": "August 2 1975"}], 
"instance of": [{"mainsnak": "Human"}], 
"country of citizenship": [{"mainsnak": "Germany"}, 
{"mainsnak": "China"}], 
"place of birth": [{"mainsnak": "Guiyang"}], 
"sport": [{"mainsnak": "Badminton"}] \end{verbatim}

\item[TML52:] Xu Huaiwen (born August 2 1975 in Guiyang Germany) is a retired Badminton from China.

\item[TT52:] Xu Huaiwen (born August 2 1975) is from Germany\$\$China.

\end{description}

\end{small}

\end{document}